# WILD ANIMAL CLASSIFIER USING CNN

Sahil Faizal[1], Sanjay Sundaresan[2]

*Student, School of Computer Science and Engineering*[1]

*Student, School of Computer Science and Engineering*[2]

*Vellore Institute of Technology, Chennai*

sahilfaizal2019@gmail.com[1]

sanjay.research3@gmail.com[2]

**Abstract:** Classification and identification of wild animals for tracking and protection purposes has become increasingly important with the deterioration of the environment, and technology is the agent of change which augments this process with novel solutions. Computer vision is one such technology which uses the abilities of artificial intelligence and machine learning models on visual inputs. Convolution neural networks (CNNs) have multiple layers which have different weights for the purpose of prediction of a particular input. The precedent for classification, however, is set by the image processing techniques which provide nearly ideal input images that produce optimal results. Image segmentation is one such widely used image processing method which provides a clear demarcation of the areas of interest in the image, be it regions or objects. The Efficiency of CNN can be related to the preprocessing done before training. Further, it is a well-established fact that heterogeneity in image sources is detrimental to the performance of CNNs. Thus, the added functionality of heterogeneity elimination is performed by the image processing techniques, introducing a level of consistency that sets the tone for the excellent feature extraction and eventually in classification.

## I. INTRODUCTION

The expansion of urban areas in modern times has resulted in widespread displacement of habitats in forested areas. As a result, wild animals are forced to venture into the human settlements that often infringe on their routine activities. More often than not, food is the primary motivator for such peregrinations. It is at this point that there is tangible danger to any humans that inadvertently cross the path of these animals when they are at their most ferocious predispositions. Hence, a need arises for the detection of wild animals at the border of human settlements close to wild habitats.

A robust, reliable and effective preemptive warning mechanism would drastically eliminate risk of fatal human-animal conflict, both in the interest of protecting human lives and avoiding loss of endangered animals. Moreover, such a system would also be useful in wildlife sanctuaries and biosphere reserves to monitor the movement of animals at the border areas of such establishments which have often proved difficult to control. The usage of technology and robust cameras is not an alien concept in most major biosphere reserves and national parks around the world. Although there has been a considerable amount of progress, software-based tools have not been explored to a satisfactory extent in these use cases.

Computer vision has the ability to transform the tracking and monitoring process with the accuracy that its components and supporting techniques provide. The automation-augmented reduction of man-hours invested in searching for and tracking wild animals is perhaps the biggest potential boon that computer vision can provide. The pre-processing involved in the application of computer vision algorithms is often under-documented although it plays a key role in the success of the algorithm. A deep understanding of the nature of the inputs is necessary to make appropriate changes at crucial junctures of processing to meet the often-convoluted criteria required by complicated deep learning algorithms. Transforming the images is invariably necessitated due to the erratic nature of real-world data feeds. The absence of an artificial synthesis element in the generation of inputs via raw camera stills adds to the intricacies involved in the image processing component.

In this paper, we are performing classification of wild animals listed in the IUCN Red List of Threatened Species. The dataset for the same has been sourced from Kaggle [1]. The images are captured in diverse lighting conditions and environments thereby making our algorithm more capable of real world deployment in conservation setups.

## II. RELATED WORKS

Wildlife protection has had a significant number of approaches which integrate myriad technology stacks to solve niche issues.

A system that combines deep learning classification with dynamic background modelling to evolve a swift and precise method for human and animal detection from highly cluttered camera trap pictures. Background modelling helps generate region proposals for foreground objects, which are then classified using the DCNN, resulting in improved efficiency and increased accuracy. The proposed system achieves 82% accuracy in segmenting images into human, animal and background patches [2].

A.V Sayagavi et al. [3] used a network of cameras, connected to PIR motion sensors, so that image capture is triggered only when some movement is detected. The images captured through these cameras are processed to detect the presence of wild animals using YOLO, and if an animal is found, identify the species. Once identified, the animals are tracked for a suitable time using CSRT in order to determine their intent – such as to find whether they are moving across the village, or into it. In the latter case, alerts are generated and local authorities are notified through proper channels. The models at present can detect 5 types of animals namely (elephant, zebra, giraffe, lion and cheetah).

A comparative study on 4 different algorithms based on deep neural networks has also been proposed in [4]. Two variants of single shot multibox detector (SSD) and two variants of faster region-based CNN (Faster R-CNN) have been compared. Two different activation functions are also compared. The SSD variants outperform all faster R-CNN variants and provide more precise detection compared to the latter.

Non-intrusive monitoring of animals was explored by A.G Villa et al. [5]. They generated huge volumes of data as they made use of multiple camera trapping networks. To analyse the data, they used a very deep CNN framework and chose 26 species from the Snapshot Serengeti (SSe) dataset. The proposed model achieved an accuracy of 88.9%. A comparison with other techniques was also carried out which showcased that their model outperformed previous approaches.

An automated wildlife monitoring system which leverages state-of-the-art deep CNN architecture has been devised [6]. The model achieved 90.4% accuracy for 3 animal classes. A single labelled dataset was used for training purposes, and a focus is placed on filtering animal images.

A comparative study between the bag of visual words and deep learning CNN techniques for wild animal recognition [7]. The comparison is done for grey level as well as colour information. The features extracted by the BOW models were combined with a regularised L2 support vector machine for classification. This study suggests that there is a clear performance gulf between CNN methods and BOW.

In [8] YOLOv3, a CNN architecture as a pre-trained model through transfer learning technique is used. Fine tuning was subsequently performed using an amalgamation of self-shot and crowdsourced images. The model locates the object detected and adds a bounding box upon it.

### III. PROPOSED METHODOLOGY

The intrinsic nature of any classification technique is the ability to accurately identify the major features of the target that it aims to predict. It is a well-documented fact that Deep Learning models and frameworks provide enhanced accuracy when the inputs have decreased source-induced heterogeneity. Since raw real-world feeds do not generate ideal or optimal images for classification, the onus is on the application of image processing techniques to act as the liaison between the classifier and the inputs. Some of the most influential challenges to classifier performance with regards to light intensity, unavailability of high-quality night-vision training images, noisy or element-rich backgrounds of captured images and luminance problems because of shadow effect need to be mitigated. The goal is to enable the classifier to extract features optimally from the images in order to assign a class to them with minimal loss and maximum accuracy. Ergo, the preprocessing needs to be performed specifically to ensure that the features are as distinctly visible as possible. The proposed sequence of operations on input images involves:

1. K-Means based segmentation via OpenCV, with a K-Value of 3
2. Noise removal in segmented images in order to increase the contrast between regions, ensuring the highest level of distinction.
3. Data Augmentation by flipping and rotating the noise-free segmented images to increase the diversity of the training set with respect to orientation.
4. Feeding the processed images to the InceptionResNetV2 classifier which predicts the class of the animal.
5. Using the convolutional layers of pre-trained neural networks with existing pre-trained values which were obtained from highly diverse bulk training samples.
6. Completely customising the fully connected layers, keeping the entire model open for training and updating weights.

The usage of K value of 3 is motivated primarily by the need to minimise induced noise in the input images and subsequently avoid adding to the existing noise. A higher K-value would theoretically be useful for classification, but it was observed to result in incoherent boundaries and significant overlap between regions. The necessity for noise removal is once again dictated by the need to optimise the images for feature extraction. Since InceptionResNetV2 classifier is CNN-based with 164 layers, a larger and more imperatively diverse training set is ideal.

Diversity with respect to the number of samples distributed across classes exists in standardised datasets, but orientation-based diversity has to be artificially introduced, thus propagating the need for data augmentation. At this juncture in the processing pipeline, fine tuning is introduced via transfer learning technique, primarily to boost the performance of the classifier and bridge the gap to its optimal accuracy. As the pre-trained weights of InceptionResNetV2 are proven to be effective in image-classification use-cases, the convolutional layers of InceptionResNetV2 are set with pre-trained weights. In the interest of flexibility, however, the fully connected layers are built according to the specifics of the training set. The usage of this model-building approach is the ideal balance between accuracy and flexibility. The proposed approach also differs from those documented in section II with regards to the number of animal classes under consideration. The 10 classes of animals are as listed in Figure 1 below:

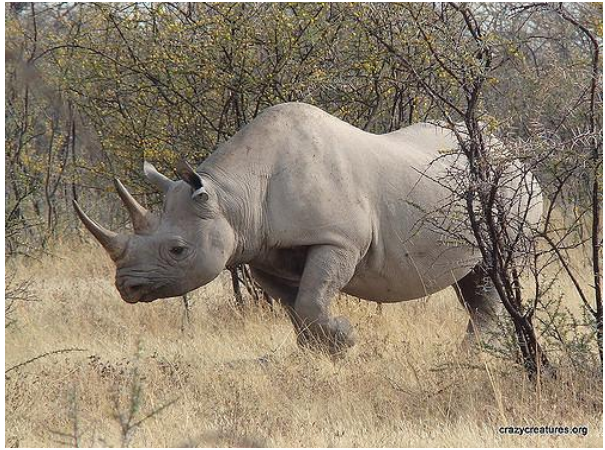

Fig. 1. Animal Classes under consideration

Additionally, class imbalance is also eliminated by the data augmentation performed in step 3, where low sample sizes for classes are mitigated if present.

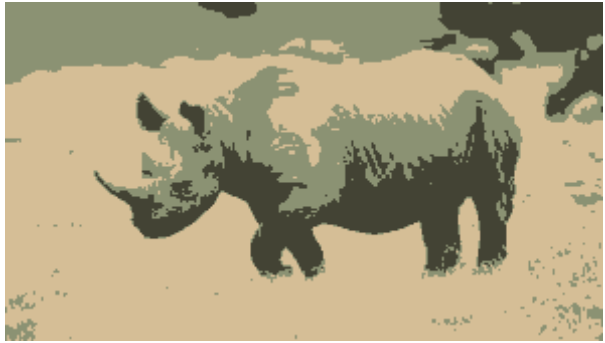

Fig. 2. Sample Input Image

Fig. 3. Corresponding Segmented Image

## IV. PROPOSED SYSTEM DESIGN

The system can be divided into four parts - data collection, preprocessing, re-sampling and model building.

*A. Data Collection: -*

The input images were sources from a curated dataset consisting of 6484 images collected from a Kaggle dataset on wild animals.

*B. Pre-Processing:-*

It was found that these images had a lot of distortion and noise in them so removing them was a necessity for the CNN algorithm to work well. After experimenting with various noise removal and enhancement techniques it was found that K-Means segmentation was successful enough to bring in a differentiating factor between the images as it was able to remove the background of the images leaving behind the animals in the images.

*C. Re-sampling: -*

Once this stage was done, it was noticed that significant class imbalance existed between the classes, hence some data samples were synthesised from original ones through rotation, flipping and zooming to increase sample count.

*D. Model Building: -*

Once the images were deemed fit for undergoing training, different models were built including VGG19, VGG16, InceptionV3, MobileNetV2, MobileNetV3, InceptionRestNetV2 to gauge the potential of fine-tuning technique to classify images into multiple classes based on its past experience of undergoing training with ImageNet dataset having 1 million samples. Eventually it was found that InceptionRestNetV2 had the best results after training for 40 epochs. Data Augmentation was used before feeding data into the model to bring in diversity in the dataset. The technique of Fine Tuning involves the usage of initial weights used in CNN layers as the best ones from the ImageNet training. The fully connected layers are custom-built as per our need and the entire model is subjected to weight changes by backpropagation during the trading phase. In the fully connected layer, the first and second layers have 512, 256 neurons while the third(final) has 10 with softmax activation, Adam optimizer and Categorical cross entropy loss. When overfitting occurred, between the first and second layer, a dropout layer was added to cut off 25% of the connections in between them. The evaluation of the model performance is done on the basis of accuracy obtained after the building process. The result is a proof of the fact that transfer learning [9] based models have superior performance when a limited amount of data is available for training.

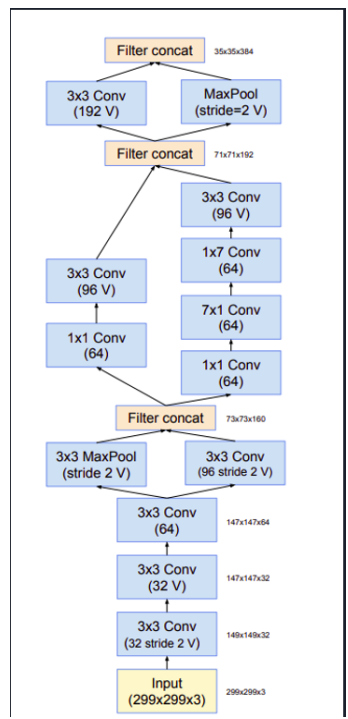

Fig. 4. Model Architecture

V. **EXPERIMENTATION AND RESULTS**

The proposed deep neural network based model made by the technique of fine tuning of the InceptionResNet-V2 network has been trained using cloud computing resources of Google Colab. For the purpose of training a dedicated NVIDIA Tesla P100 GPU with 16 GB VRAM, 25 GB memory(RAM) and 167 GB disk storage was allotted, an estimated time of 40 minutes was taken for the training purpose. The allocated VM(Virtual Machine) was having 2.3GHz Intel Xeon(R) Processor, which is used for the processing and inference. Before training all images were resized to 299 x 299 pixels in order to maintain uniformity. The resulting model is a multi-class classifier, which can classify the wild animals into 10 classes. The model was trained for 40 epochs, which has resulted in better results. The Adam optimizer with initial learning rate of 0.0001 for aiding backpropagation function of neural networks. The graph in Fig 5 shows Accuracy vs Epochs ,while in Fig 6 the plot depicts Loss vs Epochs. The details regarding the accuracy of the model upon testing with a standardised image set, a randomly generated subset of the source dataset is shown in Table I. It can be noted from the plots that the testing accuracy of the model started degrading and the testing loss started increasing after 20 epochs. Also the results obtained from the experimentation have been validated through K-Fold Cross Validation[10], in which the data is split into 5 folds and each individual fold is given as input to the model and the corresponding results are stored. Through the validation the generalisation capability of our model is also proven as the model gets exposed to unseen data as well which are unrelated to each other.

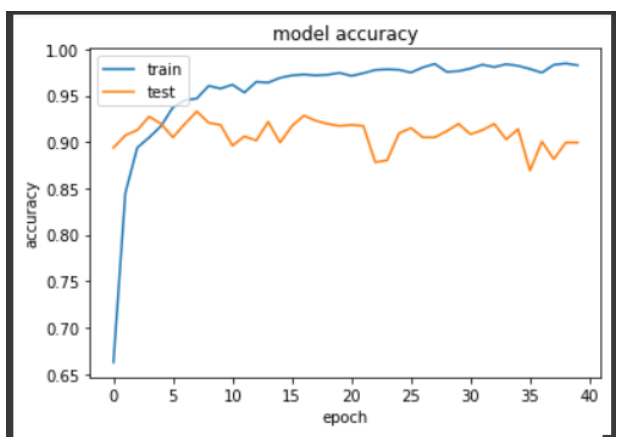

Fig. 5. Accuracy vs Epoch Curve

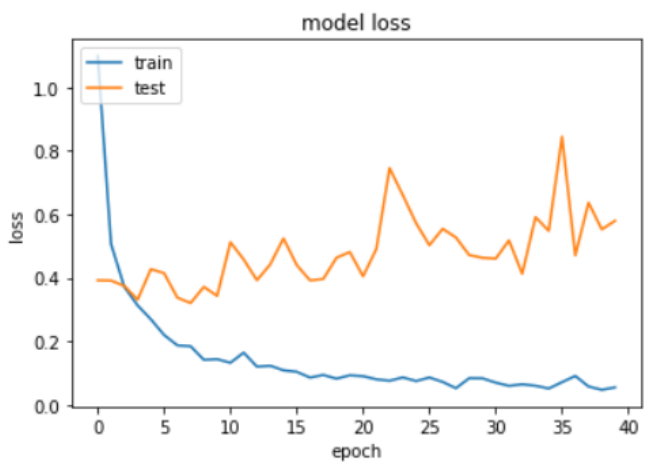

Fig. 6. Loss vs Epoch Curve

TABLE I
Accuracy Data

| Model Training Accuracy | Model Testing Accuracy | Number of Epochs |
|---|---|---|
| 95% | 91% | 40 |

## VI. CONCLUSION

Although the obtained accuracy is satisfactory for real world use cases, robust detection of wild animals exclusively at night time when there is no natural light, is potentially the most challenging yet impactful expansion. The constraints involved in processing images from night vision cameras are highly detrimental to the application of computer vision models, as the distinction between features plays the central role in the working of the model. However, it is possible to process the images generated during night hours in order to incorporate some contrast and distinction between regions of interest in the image. The availability of standardised night-vision camera datasets with a large number of animals would be the ideal catalyst for this potential expansion. Also various other deep learning based architectures and techniques can be performed.

## ACKNOWLEDGMENT

We wish to place on record our sincere gratitude to our course faculty for encouraging us to be at our innovative best by guiding us towards a unique project that seeks to be different from the work done hitherto. We also take this opportunity to thank our friends, family and peers for providing the appropriate constructive criticism at different times during the development of this project, thus enabling us to maximise the efficiency of our concept application to fulfil our potential to the best of our abilities.

## BIOGRAPHY

Sahil Faizal, is a Final-Year Student pursuing Bachelors in Computer Science Engineering at Vellore Institute of Technology, Chennai. Currently he is working in collaboration with NTU, Singapore on a research project. During undergrad studies he has worked on various research based and academic projects in the field of Deep Learning and Computer Vision. He is also the recipient of the MITACS Globalink Research Award for pursuing research based work at Dalhousie University Canada. He is keen to pursue higher studies in the field of computer science with concentration in Artificial Intelligence to bring positive changes in the lives of people.

Sanjay Sundaresan, is a Final Year undergraduate student pursuing Bachelors in Computer Science and Engineering at Vellore Institute of Technology, Chennai. As part of the project-oriented academic curriculum, he has undertaken multiple projects which use research as the foundation for impactful large-scale solutions. He believes that leveraging technology to offer holistic and eclectic solutions needs to be the primary focus, right from the grassroots level. He is eager to probe the domains of Cloud Computing, Artificial Intelligence and Internet of Things at a deeper level in a research context.